\documentclass[sigconf, screen]{acmart}
\AtBeginDocument{%
  }

\acmConference[ACM MM '26]{Proceedings of the 34th ACM International Conference on Multimedia}{November 10--14, 2026}{Rio de Janeiro, Brazil}

\usepackage{amsmath}
\usepackage{booktabs} 
\usepackage{multirow} 
\usepackage{graphicx}

\usepackage{multirow}    
\usepackage{graphicx}    
\usepackage{array}       
\usepackage{booktabs}    

\usepackage{xcolor,pifont}
\definecolor{darkgreen}{RGB}{0,100,0} 

\usepackage{amsmath,amsthm}
\usepackage[most]{tcolorbox} 
\usepackage{enumitem} 

\usepackage{longtable}   
\usepackage{tabularx}    
\usepackage{ltxtable}    
\usepackage{enumitem}    
\usepackage[table]{xcolor} 
\definecolor{closedbg}{RGB}{235,245,255} 
\definecolor{openbg}{RGB}{245,235,255}   
\definecolor{humanbg}{RGB}{235,255,235}  

\tcbset{
  mytheorem/.style={
    enhanced,
    colback=blue!5,        
    colframe=blue!60!black,
    coltitle=black,        
    fonttitle=\bfseries,   
    boxrule=0.8pt,         
    arc=2mm,               
    left=5pt,right=5pt,top=5pt,bottom=5pt, 
  }
}

\begin{document}

\title{Do MLLMs Really Understand Space?  A Mathematical Spatial Reasoning Evaluation}

\settopmatter{authorsperrow=4}

\author{Shuo Lu}
\authornote{Equal contribution.}
\affiliation{%
  \institution{NLPR \& MAIS, CASIA}
  \city{Beijing}
  \country{China}}

\author{Jianjie Cheng}
\authornotemark[1]
\affiliation{%
  \institution{Meituan Inc.}
  \city{Beijing}
  \country{China}}

\author{Yinuo Xu}
\authornotemark[1]
\author{Yongcan Yu}
\authornotemark[1]
\affiliation{%
  \institution{NLPR \& MAIS, CASIA}
  \city{Beijing}
  \country{China}}

\author{Lijun Sheng}
\author{Peijie Wang}
\affiliation{%
  \institution{NLPR \& MAIS, CASIA}
  \city{Beijing}
  \country{China}}

\author{Siru Jiang}
\author{Yongguan Hu}
\affiliation{%
  \institution{NLPR \& MAIS, CASIA}
  \city{Beijing}
  \country{China}}

\author{Run Ling}
\affiliation{%
  \institution{Northeastern University}
  \city{Shenyang}
  \country{China}}


\author{Yihua Shao}
\affiliation{%
  \institution{CASIA}
  \city{Beijing}
  \country{China}}

\author{Ao Ma}
\author{Wei Feng}
\affiliation{%
  \institution{UCAS}
  \city{Beijing}
  \country{China}}

\author{Lingxiao He}
\author{Meng Wang}
\affiliation{%
  \institution{Meituan Inc.}
  \city{Beijing}
  \country{China}}

\author{Qianlong Xie}
\author{Xingxing Wang}
\affiliation{%
  \institution{Meituan Inc.}
  \city{Beijing}
  \country{China}}

\author{Nicu Sebe}
\affiliation{%
  \institution{University of Trento}
  \city{Trento}
  \country{Italy}}

\author{Ran He}
\author{Jian Liang}
\authornote{Corresponding author.}
\email{liangjian92@gmail.com}
\affiliation{%
  \institution{NLPR, CASIA; UCAS}
  \city{Beijing}
  \country{China}}

\renewcommand{\shortauthors}{Lu et al.}


\begin{abstract}
Multimodal large language models (MLLMs) have achieved strong
performance on perception-oriented tasks, yet their ability to perform
mathematical spatial reasoning, defined as the capacity to parse and
manipulate two- and three-dimensional relations, remains unclear.
Humans easily solve textbook-style spatial reasoning problems with
over 95\% accuracy, but we find that most leading MLLMs fail to reach
even 60\% on the same tasks. This striking gap highlights spatial
reasoning as a fundamental weakness of current models.
To investigate this gap, we present \emph{MathSpatial}, the first
large-scale and systematic dataset resource dedicated to mathematical
spatial reasoning in MLLMs.
\emph{MathSpatial} provides two complementary subsets: (i)~\emph{MathSpatial-Bench}, a rigorously curated evaluation set of 2{,}000 problems spanning 3 categories and 11 subtypes, designed to isolate spatial reasoning from perceptual noise; and (ii)~\emph{MathSpatial-Corpus}, a training set of 8{,}000 problems equipped with verified solutions and structured reasoning traces. All problems are sourced from authentic educational materials and undergo multi-stage quality control including deduplication, geometric consistency checking, and cross-validated solution verification.
Benchmarking 16 leading MLLMs on \emph{MathSpatial-Bench} reveals that spatial reasoning remains a fundamental bottleneck: even GPT-5 lags behind human performance by over 35 percentage points, with particularly poor results on abstract deduction tasks. We further show that training on \emph{MathSpatial-Corpus} yields consistent improvements across model families, demonstrating the dataset's practical value for advancing spatial reasoning capabilities. \emph{MathSpatial} is publicly available at \url{https://shuolucs.github.io/MathSpatial}.
\end{abstract}


\maketitle

\section{Introduction}
\label{sec:introduction}

\begin{figure*}[t!]
  \centering
  \includegraphics[width=0.9\linewidth]{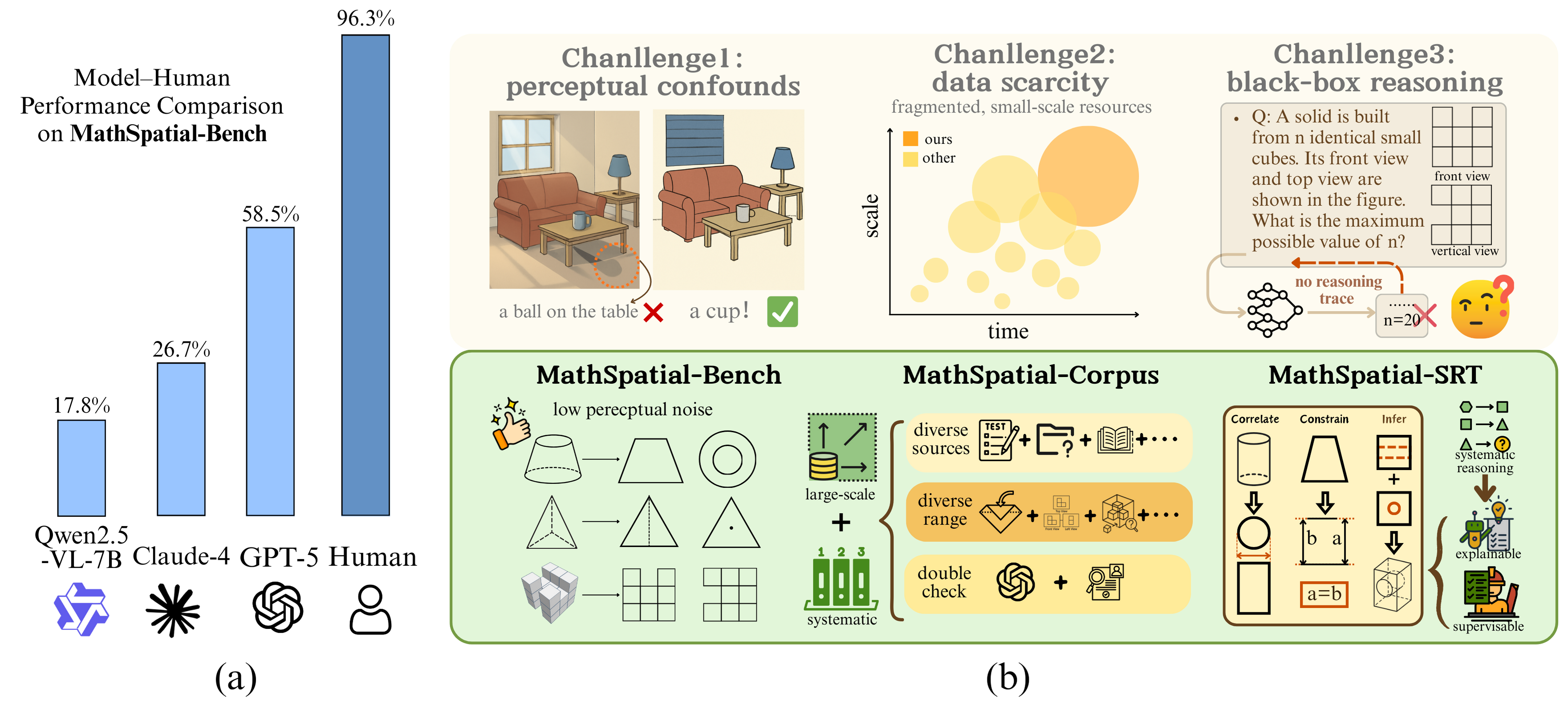}
  \caption{Left: On \emph{MathSpatial-Bench}, humans achieve over 95\% accuracy while most MLLMs remain below 60\%. Right: Three core challenges of spatial reasoning and the design of \emph{MathSpatial} to address them.}
  \label{fig:eduspacial_0923}
\end{figure*}

Multimodal large language models (MLLMs) have advanced rapidly on perception-oriented tasks~\citep{sarto2025image,kuang2025natural,ye2025llavaction}.
However, whether MLLMs can truly perform spatial reasoning, the ability to parse and manipulate 2D and 3D relationships, remains an open question.
This ability is central to higher-level cognition and underpins applications such as robotic manipulation~\citep{yuan2025seeing}, autonomous driving~\citep{tian2025nuscenes}, and embodied intelligence~\citep{feng2025survey}, 
yet we find current models frequently fail on problems that humans solve with ease, as shown in Figure~\ref{fig:eduspacial_0923}(a).

Existing research~\citep{zha2025enable,tang2025lego} on spatial reasoning has made notable progress, yet current resources suffer from three key limitations (Figure~\ref{fig:eduspacial_0923}b).
First, many benchmarks embed tasks in visually complex scenes~\citep{ma2024_3dsrbench,wang2024spatialeval,jia2025omnispatial}, where rendered details introduce perceptual confounds that lead to errors prior to reasoning, obscuring the clear analysis of spatial reasoning abilities.
Second, the field has not provided large-scale, high-quality training corpora for spatial reasoning, as shown in Table~\ref{tab:cmp_dataset}, creating a data bottleneck that limits systematic model improvement.
Third, most methods treat spatial reasoning as an end-to-end black-box mapping~\citep{chen2024spatialvlmendowingvisionlanguagemodels,ray2025satdynamicspatialaptitude}, without interpretable intermediate supervision, making it difficult to diagnose failure modes or provide targeted improvements.

Building on these insights, we introduce \emph{MathSpatial}, a large-scale dataset and benchmark ecosystem for mathematical spatial reasoning in MLLMs. Grounded in simple yet diagnostic mathematical spatial problems, \emph{MathSpatial} advances the field along three dimensions: \emph{rigorous evaluation} through a carefully curated benchmark, \emph{large-scale training data} with verified solutions, and \emph{structured annotations} that support interpretable reasoning, providing a systematic foundation for both research and model development.

To mitigate \emph{perceptual confounds}, \emph{MathSpatial-Bench} provides 2{,}000 evaluation problems using clean geometric diagrams with minimal background and texture, isolating reasoning from visual noise. It spans three categories---Holistic Recognition, Generative Inference, and Abstract Deduction---across 11 subtypes, enabling fine-grained diagnostic analysis. On this benchmark, humans achieve 95\%+ accuracy while state-of-the-art MLLMs struggle below 60\%.
To address \emph{data scarcity}, \emph{MathSpatial-Corpus} offers 8{,}000 training problems sourced from textbooks and exam banks spanning primary to high school levels, covering tasks from three-view recognition to geometric property calculation, each with human-verified solutions.
To enable \emph{interpretable reasoning}, each problem is further annotated with structured reasoning traces composed of three atomic operations (\textsc{Correlate}, \textsc{Constrain}, and \textsc{Infer}) providing explicit intermediate supervision beyond black-box training.
All data undergoes multi-stage quality control including geometric consistency checking, cross-split deduplication, and solution verification.

\begin{table}[ht]
\centering
\scriptsize 
\caption{Comparison with existing spatial/geometry reasoning benchmarks. \textit{Bilingual}: multi-language support; \textit{Spatial Focus}: emphasis on spatial reasoning over perception-heavy tasks; \textit{Train Set}: availability of training data.}
\renewcommand{\arraystretch}{1.25} 
\resizebox{\columnwidth}{!}{ 
\begin{tabular}{llllll} 
\hline
Dataset & \#Tasks & \#Samples & Bilingual & Spatial Focus & Train Set \\
\hline
EmbSpatial-Bench~\citep{du2024embspatial} & 6 & 3,640 & \multicolumn{1}{c}{{\color{red}\ding{55}}} & \multicolumn{1}{c}{{\color{red}\ding{55}}} & \multicolumn{1}{c}{{\color{darkgreen}\ding{51}}} \\
Space3D-Bench~\citep{szymanska2024space3d} & 6 & 211 & \multicolumn{1}{c}{{\color{red}\ding{55}}} & \multicolumn{1}{c}{{\color{red}\ding{55}}} & \multicolumn{1}{c}{{\color{red}\ding{55}}} \\
SpatialRGPT-Bench~\citep{cheng2024spatialrgpt} & 12 & 1,406 & \multicolumn{1}{c}{{\color{red}\ding{55}}} & \multicolumn{1}{c}{{\color{red}\ding{55}}} & \multicolumn{1}{c}{{\color{darkgreen}\ding{51}}} \\
BLINK-Spatial~\citep{fu2024blink} & 14 & 286 & \multicolumn{1}{c}{{\color{red}\ding{55}}} & \multicolumn{1}{c}{{\color{red}\ding{55}}} & \multicolumn{1}{c}{{\color{red}\ding{55}}} \\
SpatialVLM~\citep{chen2024spatialvlmendowingvisionlanguagemodels} & 2 & 546 & \multicolumn{1}{c}{{\color{red}\ding{55}}} & \multicolumn{1}{c}{{\color{red}\ding{55}}} & \multicolumn{1}{c}{{\color{darkgreen}\ding{51}}} \\
GeoEval~\citep{zhang2024geoeval} & 3 & 5,050 & \multicolumn{1}{c}{{\color{red}\ding{55}}} & \multicolumn{1}{c}{{\color{red}\ding{55}}} & \multicolumn{1}{c}{{\color{red}\ding{55}}} \\
3DSRBench~\citep{ma2024_3dsrbench} & 4 & 6,942 & \multicolumn{1}{c}{{\color{red}\ding{55}}} & \multicolumn{1}{c}{{\color{red}\ding{55}}} & \multicolumn{1}{c}{{\color{red}\ding{55}}} \\
SOLIDGEO~\citep{wang2025solidgeo} & 8 & 3,113 & \multicolumn{1}{c}{{\color{darkgreen}\ding{51}}} & \multicolumn{1}{c}{{\color{darkgreen}\ding{51}}} & \multicolumn{1}{c}{{\color{red}\ding{55}}} \\
SpatialBot-Bench~\citep{cai2025spatialbot} & 5 & 200 & \multicolumn{1}{c}{{\color{red}\ding{55}}} & \multicolumn{1}{c}{{\color{red}\ding{55}}} & \multicolumn{1}{c}{{\color{darkgreen}\ding{51}}} \\
VSI-Bench~\citep{yang2025thinkingspacemultimodallarge} & 8 & 288 & \multicolumn{1}{c}{{\color{red}\ding{55}}} & \multicolumn{1}{c}{{\color{red}\ding{55}}} & \multicolumn{1}{c}{{\color{red}\ding{55}}} \\
OmniSpatial~\citep{jia2025omnispatial} & 50 & 1,387 & \multicolumn{1}{c}{{\color{red}\ding{55}}} & \multicolumn{1}{c}{{\color{red}\ding{55}}} & \multicolumn{1}{c}{{\color{red}\ding{55}}} \\
\rowcolor[HTML]{F2F2F2} 
\textbf{MathSpatial (Ours)} & \textbf{11} & \textbf{10,000} & \multicolumn{1}{c}{{\color{darkgreen}\ding{51}}} & \multicolumn{1}{c}{{\color{darkgreen}\ding{51}}} & \multicolumn{1}{c}{{\color{darkgreen}\ding{51}}} \\
\hline
\end{tabular}
}
\label{tab:cmp_dataset}
\end{table}

In summary, our contributions are threefold:
\begin{itemize}[nosep,leftmargin=*]
    \item We collect and release \emph{MathSpatial}, a large-scale dataset of over 10{,}000 high-quality spatial reasoning problems with multi-stage quality control. It consists of \emph{MathSpatial-Bench} (2K problems across 3 categories and 11 subtypes) for diagnostic evaluation and \emph{MathSpatial-Corpus} (8K problems with human-verified solutions) for training, representing the largest dedicated resource for mathematical spatial reasoning in MLLMs.
    \item We provide structured reasoning annotations for all problems of \emph{MathSpatial-Corpus} based on three atomic operations (\textsc{Correlate}, \textsc{Constrain}, \textsc{Infer}), offering interpretable intermediate supervision to facilitate future research on spatial reasoning.
    \item We comprehensively benchmark leading MLLMs on \emph{MathSpatial-Bench}. The results show that spatial reasoning remains a significant challenge, with even GPT-5 reaching only 60\% of human accuracy. Error analysis further identifies geometric rule violations and reasoning gaps as the dominant failure modes.
\end{itemize}

\begin{figure*}[t]
  \centering
  \includegraphics[width=0.97\linewidth]{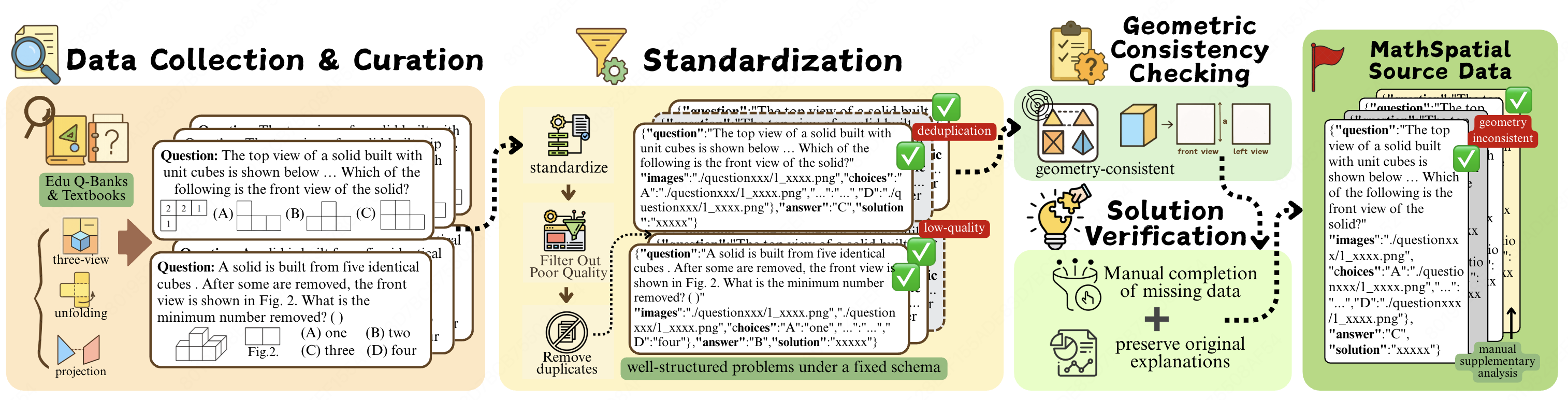}
  \caption{\emph{MathSpatial} source data construction pipeline: Data Collection and Curation $\rightarrow$ Standardization $\rightarrow$ Geometric Consistency Checking $\rightarrow$ Solution Verification.}
  \label{fig:pipeline}
\end{figure*}

\section{Related Work}
\label{sec:related_work}

\subsection{MLLM Spatial Reasoning}
Multimodal large language models (MLLMs) integrate textual and visual modalities
and have demonstrated remarkable potential across vision--language
tasks~\citep{li2023blip,chen2024expanding,bai2025qwen2,phan2025humanity}.
Recent studies indicate that MLLMs exhibit emerging spatial reasoning abilities,
such as object-relation understanding, mental rotation, and geometric
transformation~\citep{yang2025thinkingspacemultimodallarge,li2025stare,wang2025spatialviz},
typically enhanced through supervised fine-tuning on curated datasets.
For instance, SpatialVLM~\citep{chen2024spatialvlmendowingvisionlanguagemodels}
and SAT~\citep{ray2025satdynamicspatialaptitude} leverage geometric labels and
spatial rules, while MM-Spatial~\citep{daxberger2025mmspatialexploring3dspatial}
and Spatial-MLLM~\citep{wu2025spatialmllmboostingmllmcapabilities} employ
multi-view 3D scenes and large-scale corpora to strengthen cross-view reasoning.
However, these approaches rely on free-form CoT supervision and lack large-scale, systematically curated training data for mathematical spatial reasoning, a gap that \emph{MathSpatial-Corpus} directly addresses.

\subsection{Spatial Reasoning Benchmarks}
Several benchmarks have been proposed to evaluate spatial reasoning in
MLLMs~\citep{ma2024_3dsrbench,wang2024spatialeval}.
SpatialEval~\citep{wang2024spatialeval} examines spatial relations through
modality-controlled tasks; GeoEval~\citep{zhang2024geoeval} aggregates geometry
problems in text and diagram formats; 3DSRBench~\citep{ma2024_3dsrbench} tests
robustness across viewpoints; and OmniSpatial~\citep{jia2025omnispatial} covers
dynamic reasoning and perspective-taking.
However, most existing benchmarks are evaluation-only and involve
high-perception scenarios such as natural scenes and 3D
environments~\citep{yang2025mmsi,stogiannidis2025mind}, making it difficult to
assess pure reasoning ability in isolation.
\emph{MathSpatial} addresses this by providing clean geometric diagrams that eliminate perceptual confounds, paired with a large-scale training corpus, the first unified resource designed not merely to expose the spatial reasoning deficits of MLLMs, but to systematically overcome them.

\section{MathSpatial}
\label{sec:method}
\subsection{Overview}


In this section, we present \emph{MathSpatial}, a large-scale dataset for evaluating and advancing mathematical spatial reasoning. It comprises 10K problems: 2K form \emph{MathSpatial-Bench} for diagnostic evaluation and 8K form \emph{MathSpatial-Corpus} for systematic training. Beyond standard problem-solution pairs, every item in the corpus is annotated with structured reasoning traces (\emph{MathSpatial-SRT}) organized around three atomic operations (\textsc{Correlate}, \textsc{Constrain}, \textsc{Infer}), providing interpretable intermediate supervision that goes beyond conventional answer-only or free-form CoT annotations.

\begin{figure}[!ht]
  \centering
  \includegraphics[width=\linewidth]{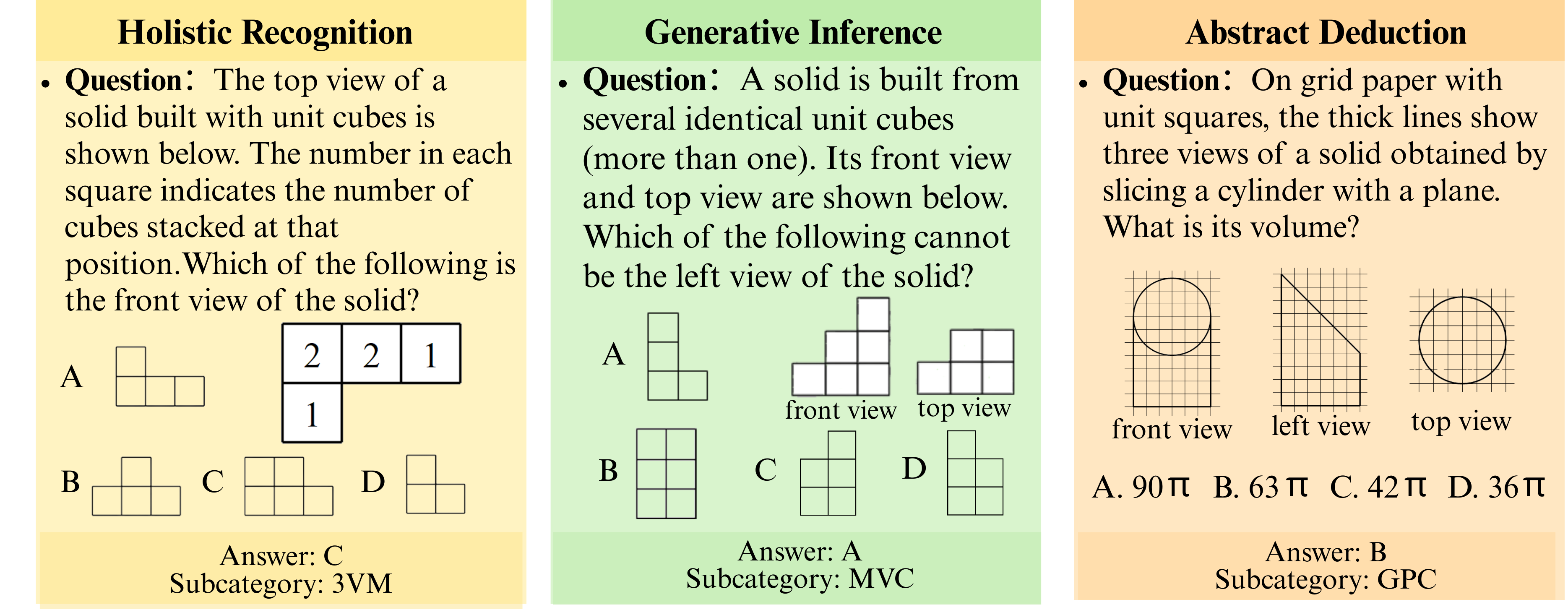}
    \caption{Selected examples demonstrating the diverse problem types across the three categories of \emph{MathSpatial}.}
  \label{fig:Mathspacial-example}
\end{figure}

\subsection{MathSpatial Data Collection}
\label{subsec:pipeline}

To construct \emph{MathSpatial}, we design a semi-automated pipeline that ensures both large scale and high quality. Problems are sourced from publicly available educational repositories and textbooks spanning primary to high school levels.\footnote{All problems are derived from public educational materials. Copyright-sensitive items are excluded to ensure compliance.} 
The tasks focus on key aspects of mathematical spatial reasoning, such as multi-view matching and unfolding/folding, among others.
The detailed data collection and curation pipeline is broken down as follows:

\begin{itemize}[nosep,leftmargin=*]
\item 
\noindent\textbf{Data Collection and Curation.}  
As illustrated in Figure~\ref{fig:pipeline}, we systematically collect spatial reasoning problems from diverse educational sources such as Baidu Wenku, Zujuan, and other online exam banks and repositories. 
To mitigate dataset-level bias, we source problems across different grades, regions, and textbook editions, focusing on questions with objective numeric or multiple-choice answers. 
These problems are inherently formalized and structured, which reduces perceptual noise. 
In total, 35,428 raw candidates were initially collected. We then executed a rigorous preliminary curation step to filter out structurally incomplete items and non-spatial queries, ultimately retaining a high-quality pool of 21,673 problems for subsequent processing.

\item 
\noindent\textbf{Preprocessing and Standardization.}  
Following the pipeline in Figure~\ref{fig:pipeline}, we standardize all problems into a unified schema containing images, questions, choices, answer, and solutions. To prevent data leakage, we enforce rigorous cross-split de-duplication using MD5 hashing, GPT-4.1 vision-aided visual similarity analysis, and semantic text filtering, followed by manual verification. Low-quality items (e.g., blurry figures or ambiguous statements) are systematically discarded. This process yields approximately 11K unique, high-quality samples. Finally, Chinese source problems are translated into English via API, with quality assured through random spot-checks.

\item 
\noindent\textbf{Geometric Consistency Checking.} As depicted in Figure~\ref{fig:pipeline}, to guarantee geometric validity, we apply rule-based verification with human-in-the-loop review. This involves checking correspondence of length–width–height across views, enforcing dashed/solid line conventions, and validating orthographic projection rules. During this stage, approximately 0.4K geometrically inconsistent items were identified and removed. All remaining items are evaluated according to a geometric QA rubric, and cases for which annotators cannot reach consensus are discarded.

\item 
\noindent\textbf{Solution Verification.} In the final stage of our pipeline (Figure~\ref{fig:pipeline}), for problems with official solutions, we retain them directly to maintain authoritative accuracy. Among all the geometrically valid items, $\sim$0.8K lacked official solutions. To address this, we recruited graduate students trained in geometry and engineering drawing to derive comprehensive solutions. Each generated solution undergoes dual-reviewer cross-validation to ensure logical completeness and reproducibility.

\item 
\noindent\textbf{Final Output.}  
Following all filtering stages, 10,000 fully verified problems remain. We partition these into MathSpatial-Bench (2K) for evaluation and MathSpatial-Corpus (8K) for training. All 10K items pass a unified pipeline ensuring problem integrity, label reliability, and strict separation between the corpus and the benchmark. As shown in Table 1, MathSpatial uniquely combines bilingual support, spatial-reasoning focus without perception-heavy noise, and large-scale training supervision, making it well suited to advance multimodal spatial reasoning research.
\end{itemize}



\textbf{Ethics and Licensing.} To ensure ethical compliance, \emph{MathSpatial} involves no human subjects or personally identifiable information, eliminating privacy risks, and strictly utilizes copyright-screened public materials. The dataset, training configurations, and leaderboard are publicly accessible \footnote{\url{https://shuolucs.github.io/MathSpatial}}. To mitigate potential translation artifacts, we employ manual spot-checks and welcome community feedback via our repository.

\subsection{MathSpatial-Bench}
\label{subsec:benchmark}

Building on the partition introduced above, we focus on \textit{MathSpatial-Bench}, the 2K-problem benchmark specifically designed to evaluate spatial reasoning in multimodal large language models. Rather than emphasizing scale, the benchmark prioritizes clarity and diagnostic value: problems are selected to minimize perceptual distractions and to isolate reasoning skills such as cross-view correspondence, geometric consistency, and deductive inference.

\begin{figure}[!ht]
    \centering
    \includegraphics[width=0.94\linewidth]{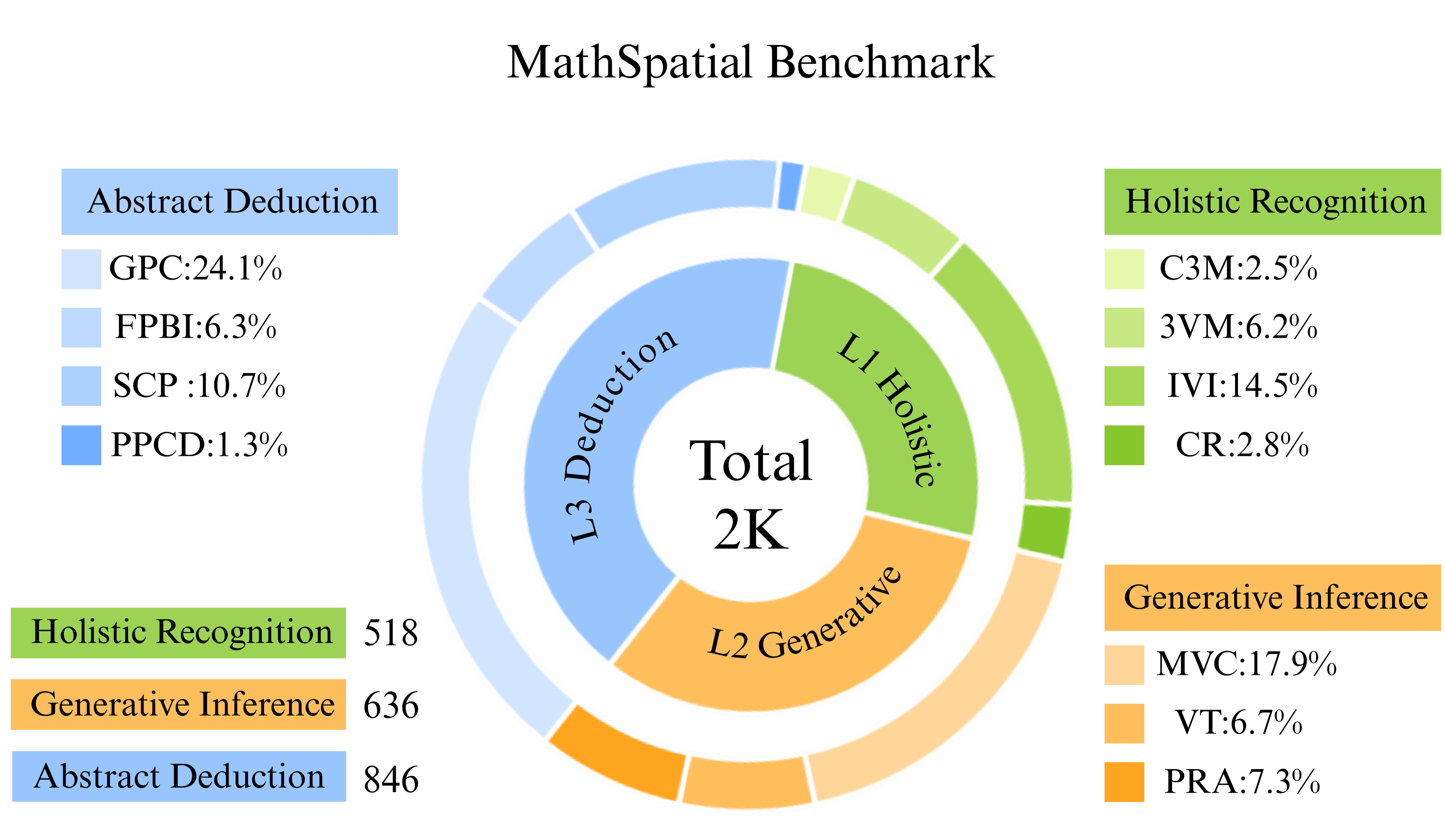}
\caption{\emph{MathSpatial-Bench} distribution and composition.}
    \label{fig:benchmark_distribution}
\end{figure}

\emph{MathSpatial-Bench} covers three major categories, further divided into eleven subcategories that reflect representative tasks in human education. The three categories are Holistic Recognition, Generative Inference, and Abstract Deduction. Figure~\ref{fig:benchmark_distribution} illustrates representative examples from each category, showing how tasks require different forms of reasoning such as multi-view correspondence, geometric consistency, and property deduction.

In total, the benchmark consists of 518 problems in Holistic Recognition, 636 in Generative Inference, and 846 in Abstract Deduction. The detailed distribution across all subcategories is shown in Figure~\ref{fig:benchmark_distribution}, where for instance Geometric Property Calculation (GPC) accounts for 24.1\% and Missing-View Completion (MVC) for 17.9\%. 
This balanced coverage enables fine-grained error analysis and a comprehensive assessment of model capabilities.

\subsection{MathSpatial-Corpus}
\label{subsec:corpus}

\begin{table*}[!ht]
\centering
\definecolor{firstbg}{HTML}{FFCCCC}  
\definecolor{secondbg}{HTML}{CCE5FF} 
\definecolor{graybg}{gray}{0.92}     

\caption{Evaluation on \emph{MathSpatial-Bench}. Accuracy (\%) per subtype and average token usage. \colorbox{firstbg}{Red} and \colorbox{secondbg}{blue} indicate the best and second-best LMMs per column. \colorbox{graybg}{Gray} indicates models fine-tuned on \emph{MathSpatial-Corpus}.}
\label{tab:comprehensive_results}
\renewcommand{\arraystretch}{1.3}
\resizebox{\textwidth}{!}{
\begin{tabular}{l|cccc|ccc|cccc|c|c}
\hline
\multirow{2}{*}{\textbf{Model}} & \multicolumn{4}{c|}{\textbf{Holistic Recognition}} & \multicolumn{3}{c|}{\textbf{Generative Inference}} & \multicolumn{4}{c|}{\textbf{Abstract Deduction}} & \multirow{2}{*}{\textbf{Overall}} & \multirow{2}{*}{\textbf{Avg. Token}} \\
\cline{2-12}
& \textbf{C3M} & \textbf{3VM} & \textbf{IVI} & \textbf{CR} & \textbf{MVC} & \textbf{VT} & \textbf{PRA} & \textbf{GPC} & \textbf{FPBI} & \textbf{SCP} & \textbf{PPCD} & & \\
\hline
\multicolumn{14}{c}{\textit{\textbf{Closed-Source Models}}} \\
\hline
GPT-5~\citep{openai2025gpt5} & 28.6 & \cellcolor{firstbg} 53.7 & \cellcolor{firstbg} 66.2 & \cellcolor{firstbg} 48.2 & \cellcolor{firstbg} 61.6 & 44.8 & \cellcolor{firstbg} 71.7 & \cellcolor{firstbg} 52.3 & \cellcolor{firstbg} 60.0 & \cellcolor{firstbg} 68.1 & \cellcolor{firstbg} 57.7 & \cellcolor{firstbg} 58.5 & 676.3 \\ 
GPT-4.1~\citep{gpt41} & 2.0 & 45.5 & 43.1 & 26.8 & 24.1 & 0.0 & 55.2 & 0.0 & 29.6 & 18.3 & 46.2 & 22.6 & 676.3 \\
GPT-4o~\citep{gpt4o} & 2.0 & 36.6 & 34.1 & 33.9 & 23.5 & 1.5 & 45.5 & 0.0 & 26.4 & 16.4 & 26.9 & 19.6 & 677.4 \\
Claude 4~\citep{claude4} & 6.1 & 33.3 & 46.2 & 21.4 & 30.8 & 1.5 & \cellcolor{secondbg} 60.7 & 0.2 & 32.0 & 41.3 & \cellcolor{secondbg} 53.8 & 26.7 & 1005.5 \\
Claude 3.7~\citep{claude37sonnet} & 2.0 & 36.6 & 39.3 & 21.4 & 21.6 & 1.5 & 57.9 & 0.0 & 23.2 & 26.3 & 34.6 & 21.5 & 885.8 \\
Claude 3.5~\citep{claude35sonnet} & 2.0 & 32.5 & 42.8 & 23.2 & 26.9 & 0.7 & 57.2 & 0.0 & 24.8 & 21.1 & 46.2 & 22.3 & 858.6 \\
Gemini-2.5-pro~\citep{gemini25_report} & \cellcolor{secondbg} 31.4 & 42.5 & 43.5 & 26.7 & \cellcolor{secondbg} 58.5 & \cellcolor{secondbg} 44.9 & 28.1 & 45.6 & \cellcolor{secondbg} 41.7 & 49.6 & 33.3 & 44.9 & 913.8 \\ 
Gemini-2.5-flash~\citep{gemini25_report} & \cellcolor{firstbg} 34.3 & 42.5 & \cellcolor{secondbg} 46.7 & 26.7 & \cellcolor{secondbg} 58.5 & \cellcolor{firstbg} 47.8 & 43.8 & \cellcolor{secondbg} 50.0 & 25.0 & \cellcolor{secondbg} 57.4 & 33.3 & \cellcolor{secondbg} 48.5 & 1115.2 \\
\hline
\multicolumn{14}{c}{\textit{\textbf{Open-Source Models}}} \\
\hline
Qwen2.5-VL-7B~\citep{bai2025qwen2} & 0.0 & 32.5 & 34.8 & 17.9 & 24.4 & 0.0 & 41.4 & 0.0 & 20.0 & 13.6 & 15.4 & 17.8 & 465.3 \\
InternVL3-8B~\citep{zhu2025internvl3} & 0.0 & 29.3 & 35.9 & 23.2 & 21.3 & 0.7 & 40.0 & 0.0 & 19.2 & 14.6 & 15.4 & 17.4 & 473.5 \\
Llama3-8B~\citep{dubey2024llama} & 10.2 & 15.4 & 20.7 & 17.9 & 15.1 & 20.9 & 23.4 & 9.3 & 13.6 & 12.2 & 7.7 & 15.0 & 785.4 \\
Qwen2.5-VL-72B~\citep{bai2025qwen2} & 2.0 & 33.3 & 35.9 & 23.2 & 26.9 & 0.0 & 47.6 & 0.0 & 20.8 & 17.8 & 19.2 & 19.7 & 497.6 \\
GLM-4.5V~\citep{hong2025glm} & 4.1 & 32.5 & 43.4 & 30.4 & 23.0 & 1.5 & 57.9 & 0.0 & 22.4 & 14.1 & 30.8 & 21.0 & 1391.1 \\
\rowcolor{graybg} MathSpatial-Qwen2.5-VL-7B  & 4.1 & 40.7 & 46.2 & 30.4 & 27.7 & 0.7 & 46.2 & 0.0 & 21.6 & 17.4 & 26.9 & 22.1 & \cellcolor{secondbg} 351.9 \\
\rowcolor{graybg} MathSpatial-InternVL3-8B & 2.0 & \cellcolor{secondbg} 50.4 & 36.2 & 33.9 & 22.7 & 11.9 & 56.6 & 3.9 & 23.2 & 15.5 & 15.4 & 22.6 & \cellcolor{firstbg} 318.3 \\
\rowcolor{graybg} MathSpatial-Llama3-8B & 16.3 & 30.1 & 24.5 & \cellcolor{secondbg} 41.1 & 20.4 & 30.6 & 27.6 & 10.2 & 20.8 & 16.4 & 7.7 & 20.3 & 397.3 \\
\hline
\multicolumn{14}{c}{\textit{\textbf{Human Performance}}} \\
\hline
Human & 100.0 & 89.4 & 97.6 & 98.2 & 96.1 & 97.0 & 95.2 & 96.5 & 94.4 & 97.7 & 96.2 & 96.3 & -- \\
\hline
\end{tabular}}
\end{table*}

\emph{MathSpatial-Corpus} provides 8{,}000 problems for training, each comprising a problem image, textual description, final answer, and detailed solution. A portion of the problems originally in Chinese is translated into English, enabling bilingual support. Together with \emph{MathSpatial-Bench}, it completes the evaluation-training loop of \emph{MathSpatial} and represents one of the largest dedicated resources for spatial reasoning.

\noindent\textbf{Structured Reasoning Annotations.}
Beyond standard problem-solution pairs, every item is annotated with structured reasoning traces organized around three atomic operations: \textsc{Correlate} (establishing cross-view correspondences between geometric entities), \textsc{Constrain} (applying geometric and projection rules to enforce consistency), and \textsc{Infer} (deducing latent attributes or final answers).
Concretely, given a problem $(x,y)$, we use GPT-4o~\citep{gpt4o} to generate operation-level traces $r=(r_1,\dots,r_T)$ under a constrained schema specifying operation type, arguments, and assertion. To ensure annotation quality, each trace is validated through a dual-role review scheme: a Reviewer agent audits every step for operation-type errors, contradictions, or missing steps, and a Checker agent rewrites the trace to correct identified issues while preserving the schema. This process detects and fixes ${\sim}$10\% of generated traces.

\noindent\textbf{Generation and training.}
Given a problem $(x,y)$, we use GPT-4o to generate operation-level traces $r=(r_1,\dots,r_T)$ under a constrained schema (operation type + arguments + assertion). We linearize $(r,y)$ into a single sequence and apply supervised fine-tuning. This strategy enables models to learn not only to predict correct answers but also to produce interpretable, verifiable reasoning traces that reflect the atomic operation structure.
To  reduce structural or logical noise, each SRT is validated through a dual-role scheme. In this process, a Reviewer agent audits every \textsc{CORR/CONS/INFER} step for operation-type errors, contradictions, or missing steps, and a Checker agent rewrites the trace to correct all identified issues while preserving the SRT schema.

\noindent\textbf{Corpus Statistics.} The corpus spans 11 subtypes with a K-12 skew: GPC dominates (65\%), while rare tasks (e.g., CR) are ${<}1\%$. It contains ${\sim}$20K images (2.6/problem) in mixed formats (46\% choice, 54\% fill-in). Crucially, all bilingual problems feature structured reasoning traces (avg.\ 5.3 steps) for robust supervision.


\section{Experiments}
\label{sec:experiments}

\subsection{Experimental Setup}
\label{subsec:setup}
We evaluate a broad range of MLLMs on \emph{MathSpatial-Bench} to assess the benchmark's diagnostic capability across model families.
Closed-source systems include GPT-5~\citep{openai2025gpt5}, GPT-4.1~\cite{gpt41}, GPT-4o~\cite{gpt4o}, Claude-3.5/3.7/Sonnet-4~\cite{claude35sonnet,claude37sonnet,claude4}, and Gemini-2.5-Pro/Flash~\cite{gemini25_report}.
Open-source baselines include Qwen2.5-VL-7B/72B~\cite{bai2025qwen2}, InternVL3-8B~\cite{zhu2025internvl3}, Llama3-8B~\cite{dubey2024llama}, and GLM-4.5V~\cite{hong2025glm}. Results for additional models (DeepSeek-VL2, Pixtral-12B, Kimi-VL, Llama-4-Maverick, Idefics3, etc.) are available on our project page.
To demonstrate the training value of \emph{MathSpatial-Corpus}, we fine-tune three open-source models (Qwen2.5-VL-7B, InternVL3-8B, Llama3-8B) on the corpus using a batch size of 128, learning rate $1\times10^{-5}$, and 5 epochs.
To establish a rigorous human upper bound, we recruited 80 students under closed-book conditions, with each problem answered by 20 independent annotators and results reported as micro-averaged accuracy. To ensure robustness, all model results are averaged over two independent runs.

\begin{figure*}
    \centering
    \includegraphics[width=0.94\linewidth]{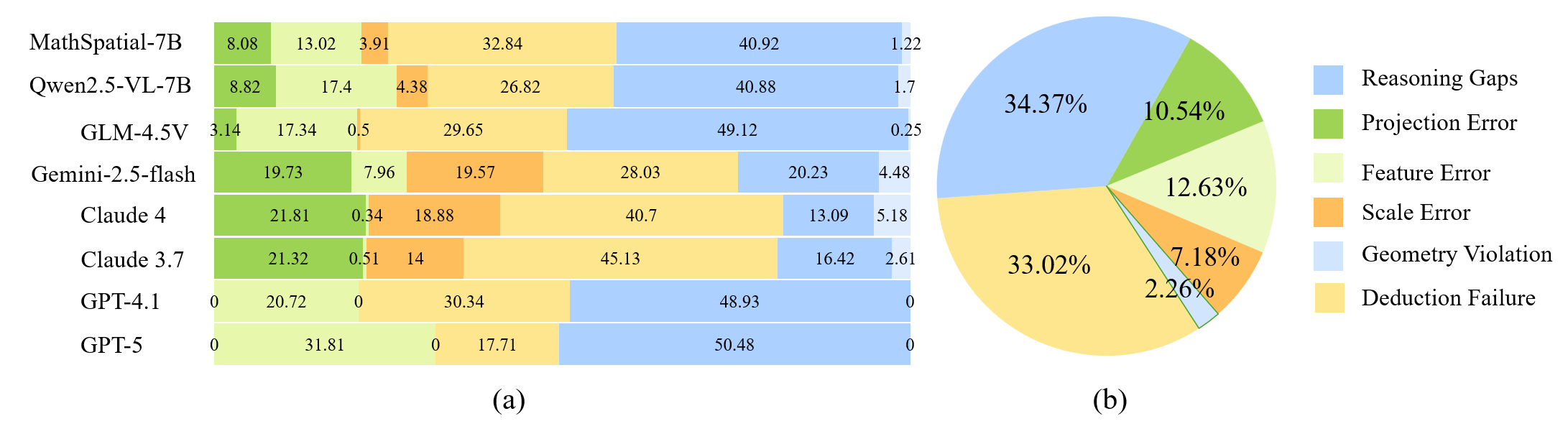}
    \caption{Fine-grained error analysis on \emph{MathSpatial-Bench}. (a) Error frequency distribution for baselines across 6 subcategories. (b) Overall error rate breakdown by failure mode.}
    \label{fig:analysis}
\end{figure*}

\subsection{Benchmark Results}
\label{subsec:comparative}

Table~\ref{tab:comprehensive_results} presents the evaluation of 16 representative models on \emph{MathSpatial-Bench}. The results reveal a striking human--model gap: humans exceed 95\% accuracy across all categories, while most MLLMs remain below 60\%.

\noindent\textbf{Closed-source models lead but remain far from human.}
GPT-5 achieves the highest accuracy (58.5\%) and dominates most subtypes, yet still reaches only 60\% of human performance (96.3\%). Gemini-2.5-flash (48.5\%) and Gemini-2.5-pro (44.9\%) follow as the second tier, showing particular strength on visual transformation tasks (VT: 47.8\% and 44.9\%) and missing-view completion (MVC: 58.5\% each). In contrast, the GPT-4 series (19.6\%--22.6\%) and Claude family (21.5\%--26.7\%) lag significantly, scoring near zero on numerically demanding tasks such as GPC (0.0\%--0.2\%) and VT (0.0\%--1.5\%).

\noindent\textbf{Open-source models lag substantially.}
Most open-source models cluster in the 15\%--21\% range. Even scaling up provides limited benefit: Qwen2.5-VL-72B (19.7\%) only marginally improves over its 7B variant (17.8\%). Multiple models score 0.0\% on GPC, indicating a complete inability to perform geometric property calculations. GLM-4.5V (21.0\%) is the strongest base open-source model on this benchmark, yet still trails most closed-source systems.

\noindent\textbf{Difficulty varies sharply across reasoning categories.}
Holistic Recognition is the most accessible: top models reach 40\%--66\% on subtypes like IVI and 3VM. Generative Inference presents a mixed picture, with MVC well above chance for strong models but VT near zero for most. Abstract Deduction is the hardest category overall; while PRA is relatively tractable (GPT-5: 71.7\%), GPC remains a near-universal failure point ($\leq$4\% for all open-source models, $\leq$52\% even for GPT-5), highlighting that multi-step geometric computation is a fundamental bottleneck.

\noindent\textbf{Training on MathSpatial-Corpus demonstrates clear dataset value.}
Fine-tuning on \emph{MathSpatial-Corpus} consistently lifts performance across all three backbone models. MathSpatial-InternVL3-8B reaches 22.6\% (vs.\ 17.4\% base), achieving the second-highest 3VM score (50.4\%) across all models. MathSpatial-Qwen2.5-VL-7B improves to 22.1\% with gains on IVI (+11.4\%), PPCD (+11.5\%), and 3VM (+8.2\%). MathSpatial-Llama3-8B shows the largest VT improvement (20.9\% $\to$ 30.6\%). Notably, all fine-tuned models also reduce token consumption by 20\%--30\% (e.g., Qwen2.5-VL-7B: 465.3 $\to$ 351.9), demonstrating that the dataset's structured annotations promote both accuracy and reasoning efficiency.

\subsection{Error Analysis}
\label{subsec:benchmark-eval}

To better understand the limitations of current MLLMs on mathematical spatial reasoning, we conduct a fine-grained error analysis on \emph{MathSpatial-Bench}. Errors are grouped into six categories: \textbf{(1)} Projection Error, where models misinterpret top, side, or front views; \textbf{(2)} Feature Error, such as omitting small components or inventing edges; \textbf{(3)} Scale Error, failing to preserve relative sizes; \textbf{(4)} Geometry Violation, where outputs break orthographic or visibility rules; \textbf{(5)} Reasoning Gaps, reflecting incomplete or inconsistent chains of thought; and \textbf{(6)} Deduction Failure, where models cannot synthesize multiple cues into a final conclusion.

Figure~\ref{fig:analysis}(a) reveals distinct weakness profiles across model families: the GPT-5/4 series is dominated by reasoning gaps, Claude models frequently violate geometric rules, Gemini-2.5-flash suffers more from projection and scale errors, while open-source models like Qwen2.5-VL-7B exhibit primarily reasoning gaps but fewer feature errors. Notably, \emph{MathSpatial-Qwen2.5-VL-7B} (fine-tuned on our corpus) reduces reasoning gaps compared to its base model, suggesting that the structured annotations in \emph{MathSpatial-Corpus} help mitigate this dominant failure mode.
As summarized in Figure~\ref{fig:analysis}(b), the majority of errors come from \textbf{reasoning gaps (34.4\%)} and \textbf{geometry violations (33.0\%)}, followed by projection (12.6\%) and feature errors (10.5\%). Scale (7.2\%) and deduction failures (2.3\%) are less frequent but still harmful.

These findings highlight two core challenges for future research: \emph{(i)} enforcing low-level geometric consistency across views, and \emph{(ii)} sustaining coherent multi-step reasoning chains. Both remain largely unsolved by current MLLMs, and the fine-grained error taxonomy provided by \emph{MathSpatial-Bench} offers a diagnostic foundation for targeted improvements.



\vspace{-7pt}
\section{Conclusion and Future Work}
\label{sec:conclusion}

We presented \emph{MathSpatial}, a large-scale, publicly available dataset for mathematical spatial reasoning in MLLMs. \emph{MathSpatial} comprises \emph{MathSpatial-Bench} (2K problems) for diagnostic evaluation across 3 categories and 11 subtypes, and \emph{MathSpatial-Corpus} (8K problems) with verified solutions and structured reasoning annotations based on three atomic operations. Comprehensive benchmarking of leading MLLMs reveals a substantial human--model gap, with humans exceeding 95\% accuracy while most models remain below 60\%. Fine-grained error analysis identifies geometric rule violations (33.0\%) and reasoning gaps (34.4\%) as the dominant failure modes, offering actionable insights for future model development. Fine-tuning on \emph{MathSpatial-Corpus} consistently improves both accuracy and efficiency across multiple model families, demonstrating the dataset's practical value for advancing spatial reasoning capabilities.
We hope \emph{MathSpatial} will accelerate progress in multimodal mathematical spatial reasoning and related downstream applications.

\noindent \textbf{Limitations and Future Work.} 
While \emph{MathSpatial} establishes a robust foundation for formalized spatial reasoning, its current scope primarily focuses on static geometric diagrams to minimize perceptual noise. In future work, we plan to expand the \emph{MathSpatial} series to encompass dynamic spatial transformations (e.g., continuous object motion) and complex 3D embodied environments. Additionally, we aim to explore advanced automated self-verification frameworks to further scale up the generation of high-fidelity reasoning traces.

\bibliographystyle{ACM-Reference-Format}
\bibliography{main}

\end{document}